\pdfoutput=1

\documentclass[11pt]{article}

\usepackage{naacl2021}

\usepackage{times}
\usepackage{latexsym}

\usepackage[T1]{fontenc}

\usepackage[utf8]{inputenc}

\usepackage{microtype}

\title{Unsupervised Transfer Learning in Multilingual Neural Machine Translation with Cross-Lingual Word Embeddings}

\author{Carlos Mullov \and Ngoc-Quan Pham \and Alexander Waibel \\
  Institute for Anthropomatics and Robotics \\
  KIT - Karlsruhe Institute of Technology, Germany \\
  \texttt{\{firstname\}.\{lastname\}@kit.edu} \\}

\usepackage{amsmath}
\usepackage{amssymb}
\usepackage{amsthm}
\usepackage{booktabs}
\usepackage{listings} 

\usepackage{rotating}
\usepackage{subcaption}
\usepackage{multicol}
\usepackage{array}
\usepackage{hyperref}
\usepackage[inline]{enumitem}

\DeclareMathAlphabet{\mathcal}{OMS}{cmsy}{m}{n}

\newcommand{\lang}{\ell}
\newcommand{\new}{\text{new}}
\newcommand{\newlang}{\lang_{\new}}

\newcommand{\baselangs}{\mathcal{L}_{\text{base}}}

\begin{document}
\maketitle

\begin{abstract}
    In this work we look into adding a new language to a multilingual NMT system in an unsupervised fashion.
    Under the utilization of pre-trained cross-lingual word embeddings we seek to exploit a language independent multilingual sentence representation to easily generalize to a new language.
    While using cross-lingual embeddings for word lookup we decode from a yet entirely unseen source language in a process we call \emph{blind decoding}.
    Blindly decoding from Portuguese using a basesystem containing several Romance languages we achieve scores of 36.4 BLEU for Portuguese-English and 12.8 BLEU for Russian-English.
    In an attempt to train the mapping from the encoder sentence representation to a new target language we use our model as an autoencoder.
    Merely training to translate from Portuguese to Portuguese while freezing the encoder we achieve 26 BLEU on English-Portuguese, and up to 28 BLEU when adding artificial noise to the input.
    Lastly we explore a more practical adaptation approach through non-iterative backtranslation, exploiting our model's ability to produce high quality translations through blind decoding.
    This yields us up to 34.6 BLEU on English-Portuguese, attaining near parity with a model adapted on real bilingual data.
\end{abstract}
%

\section{Introduction}

Neural machine translation (NMT) systems consist of an encoder, which encodes an input sentence into a latent numerical sentence representation space, and a decoder, which generates the target sentence from this representation \citep{cho-etal-2014-learning,bahdanau14}.
While under favourable circumstances neural networks bring significant improvements to translation performance, a major problem with them, however, is that they are extremely data hungry.
While for the most common language pairs large amounts of parallel data might be available, for most language pairs parallel data is extremely scarce.
As such one of the goals of multilingual NMT is to significantly reduce the amount of parallel data required for each given individual language pair.
Universal multilingual NMT \citep{johnson-etal-2017-googles,ha2016} employs a universal encoder so that the source languages share one single set of parameters.

Ideally this universal encoder would learn a language independent representation of the input sentences, e.g.\ learn to map semantically similar sentences onto similar neural sentence representations -- even across different languages.
Ongoing research in the field of zero-shot translation provided evidence that multilingual NMT models exhibit this property up to a certain extent, and that enforcing the similarity between sentence representations across different languages in turn also improves the zero-shot capabilities \citep{pham-etal-2019-improving}.
On the other hand, cross-lingual transfer learning research in NMT further showed that such multilingual systems can rapidly be extended to new languages on very little data \citep{neubig-hu-2018-rapid}.
\citet{neubig-hu-2018-rapid} furthermore show that going as far as training a basesystem on 58 source languages immensely helps the universal encoder in subsequently generalizing to new source languages in a supervised setting.
In this work we take this approach one step further and investigate the ability to translate to a new language with no parallel data for this language.
While integrating cross-lingual word embeddings into the model we provide two contributions in the field of multilingual NMT:
\begin{enumerate}
    \setlength{\itemsep}{0pt}
    \item To explore the generalization-ability of the universal encoder we translate from a yet entirely unseen language to a known language.
    \item Inspired by the recent advancements in unsupervised NMT we adapt the universal decoder to a new language through
      \begin{enumerate}[label=\emph{\alph*)}]
        \item using the NMT model as autoencoder
        \item backtranslation
      \end{enumerate}
\end{enumerate}

Using a German-English-Spanish-French-Italian multilingual basesystem we perform experiments in transfer learning to translate to and from Portuguese and Russian.
First, by simply decoding from Portuguese and Russian while using the cross-lingual embeddings for word lookup we demonstrate the effectiveness of applying the universal encoder to an unseen language, achieving reasonable $36.4$ BLEU and $12.8$ BLEU scores for Portuguese-English and Russian-English respectively.
We discovered that, when using the multilingual model as autoencoder, we can successfully adapt to the new language even without training to denoise -- simply by freezing the encoder in training.
Finally, we devise a method to jump-start the iterative backtranslation process oftentimes employed in unsupervised NMT, by exploiting model transfer learning ability and generate data by translating from an unseen language.
Our models trained on the synthetic parallel data attain translation performance close to what we achieve through supervised adaptation on real parallel data, falling short by an average of 0.8 BLEU for translation to Portuguese.

As our final contribution we experiment on continuous output NMT, which, while underperforming the softmax models in terms of BLEU scores, attains vastly superior performance in terms of training times as well as memory complexity.


\section{Multilingual Neural Machine Translation}%
\label{sec:machine_translation}
In training the multilingual NMT system we aim to estimate the probability $\mathbb{P}(Y_{\lang_{tgt}} = y \mid X_{\lang_{src}} = x)$ that the sentence $y$ in the target language $\lang_{tgt}$ is a suitable translation of the source sentence $x$ in the source language $\lang_{src}$.
We describe the distribution of the sentences in different natural languages $\lang$ through random variables $X_{\lang}$, $Y_{\lang}$.
A universal encoder maps the sentences from the different input distributions onto a single, shared latent distribution $H_{\text{enc}} = enc(X_{\lang})$.
A decoder is then tasked to model the probability distribution from this latent variable: $\mathbb{P}(Y_{\lang_{tgt}} = y \mid H_{\text{enc}} = c)$.
Identically to a regular bilingual NMT system, the multilingual translation system consists of a neural network, which, in a supervised setting, is trained to maximize the log-likelihood of a dataset of parallel sentences $\mathcal{D} = \{(x_1, y_1), \ldots{}, (x_n, y_n)\}$, where sentence pairs $(x_i, y_i)$ are translations of each other:
\[
    \min_{\vartheta \in \Theta}{
        \left\{-\sum_{i=1}^{n}{
            \log{
                \mathbb{P}\left( X_{\lang_{tgt}^i} = y_i \mid X_{\lang_{src}^i} = x_i, \vartheta \right)
            }
        }\right\}
    }
\]%
Note that the only difference to bilingual translation is the nature of the input and output distribution.
Thus, the only practical difference in the training a multilingual system is that instead of using a bilingual training corpus, $\mathcal{D}$ consists of a concatenation of several parallel corpora.

It is a desirable quality for the universal encoder to learn a language independent latent representation.
Such language independence may be expressed as a low variance $\mathbb{V}\{H_{\text{enc}}\}$ in the latent encoding space, as well as a high cross-lingual covariance $\mathbb{C}\{enc(X_{\lang}), enc(X_{\lang'})\}$.


\section{Related Work}
\citet{qi-etal-2018-pre} looked into the effectiveness of using pre-trained embeddings for NMT, asking whether the alignment of the embedding vectors into a shared embedding space helps in NMT and coming to the conclusion that it is helpful in a multilingual setting.
Similarly, \citet{sen-etal-2019-multilingual} carried out unsupervised multilingual NMT with the utilization of cross-lingual word embeddings.
While their approach is similar to ours, their focus lies in improving unsupervised NMT, while we mainly look into multilingual transfer learning.
\citet{neubig-hu-2018-rapid} concerned about the possibility of rapidly extending a multilingual NMT model by an unseen low-resource language.
They compare between bilingual training, multilingual training alongside a similar source language and multilingual pre-training with as many languages as possible.
They came to the conclusion that pre-training in a highly multilingual setting -- this in their case is a system with 58 source languages -- significantly improves the ability to learn the low-resourced language.
Similar to our findings, they also achieved significant performance on a yet entirely unseen language.
They, however, do not use cross-lingual embeddings, but rely on bilingual data to teach their model the cross-lingual word correlations.
\citet{kim-etal-2019-effective} conducted unsupervised cross-lingual transfer learning in NMT, transferring the parameters learned on a high resource language pair and to a new source language.
Similar to our work they therefore employed cross-lingual word embeddings, aligning the new language embeddings to the parent model source embeddings.
Their work focuses on adaptation on the source-side, furthermore in a bilingual setting.
\citet{escolano-etal-2019-bilingual} devised an approach to multilingual NMT with independent encoders and decoders that allows for zero-shot translation, as well as the addition of new languages.
\citet{siddhant-etal-2020-leveraging} looked into leveraging monolingual data for multilingual NMT.
Amongst other things they extended the model by a new language through masked denoising autoencoding.
While they did not perform backtranslation they suggest that their scenario presents a promising avenue for jump-starting the backtranslation process, which our work demonstrates to indeed be the case.

Our work is also closely related to unsupervised NMT \citep{lample2017unsupervised,lample-etal-2018-phrase,artetxe2017,artetxe-etal-2019-effective}.
\citet{lample2017unsupervised} trained an unsupervised NMT system which translates between two languages $\ell_1$ and $\ell_2$, while employing universal encoder and decoder for the simultaneous translation in both directions.
They iteratively trained the NMT model to generate $\ell_1$ and $\ell_2$ sentences through denoising autoencoding, while simultaneously training to maximize the similarity between the encoder sentence representation for $\ell_1$ and $\ell_2$ sentences.
Through unsupervised means they thus induce a shared sentence representation, essentially bootstrapping a multilingual NMT system from monolingual data.
We, on the other hand, consider a more realistic scenario where we have enough data from high resource language pairs to train a supervised multilingual NMT system, on which we then perform unsupervised transfer learning.


\section{Approach}
\begin{figure}[tpb]
    \centering
    \includegraphics[width=0.96\linewidth]{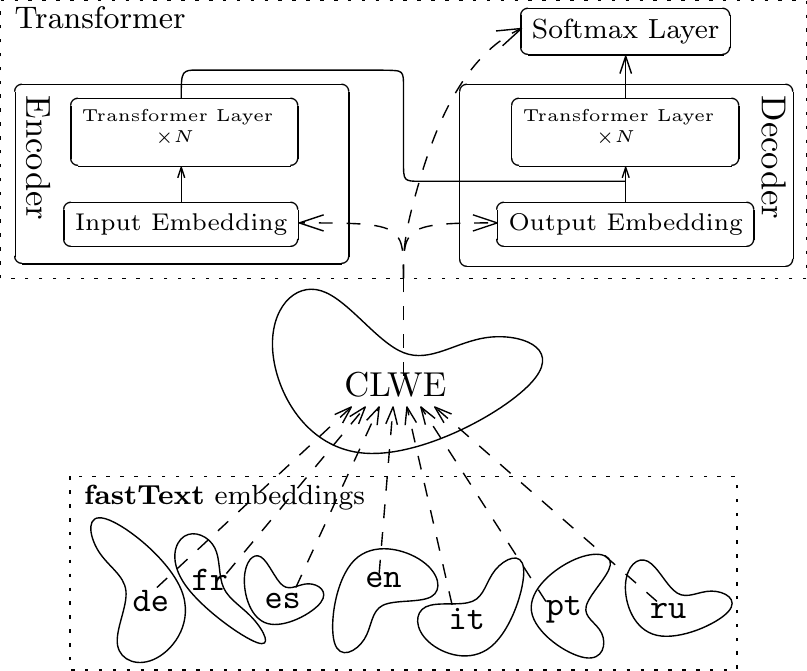}
    \caption{Our NMT architecture consists of a Transformer model with pre-trained cross-lingual word embeddings for embedding layers.
             The embedding vectors are obtained through alignment of monolingual \emph{fastText} embeddings for each language into a common embedding space.}%
    \label{fig:clwe-architecture}
\end{figure}

We propose a three-step approach to the cross-lingual transfer learning task
\begin{enumerate}
    \setlength{\itemsep}{0pt}
    \item train the cross-lingual word embeddings on monolingual data for each of the involved languages
    \item train the multilingual basesystem on parallel data for the set of base languages $\baselangs$
    \item add a new language $\newlang \notin \baselangs$ to the system on monolingual data only
\end{enumerate}
Figure \ref{fig:clwe-architecture} describes the resulting NMT architecture.%

\subsection{Cross-Lingual Word Embeddings}
\label{approach:cross_lingual_word_embeddings}
In conventional training of a universal multilingual NMT system the $\newlang$ word vectors are randomly initialized.
The model then learns in training to represent its words in a shared multilingual embedding space, by learning cross-lingual word correlations from the parallel data.
In a monolingual data only setting we aim at achieving an equivalent result through cross-lingual word embeddings.
These word embeddings are then integrated into our translation model by supplying in the form of pre-trained embeddings.

For each language $\lang \in \baselangs \cup \{\newlang\}$ we use pre-trained monolingual word embeddings learned with \emph{fastText} \citep{bojanowski-etal-2017-enriching}, which we then align into one common space through cross-lingual embedding alignment \citep{joulin-etal-2018-loss}.
For a common alignment we choose to align into a \emph{single hub space} \citep[SHS][]{heyman-etal-2019-learning}, e.g.\ we pick one of our base languages as a \emph{pivot} and align each of the embedding spaces $E_{\lang}$ to $E_{\text{pivot}}$.

\subsection{Adding a New Language}
In this work we present three methods to add a new language to a multilingual basesystem.
In the order of increasing level of supervision, they are as follows:
\begin{enumerate}
    \setlength{\itemsep}{0pt}
    \item \emph{blind} decoding
    \item (denoising) autoencoding
    \item backtranslation
\end{enumerate}
The first experiment serves to answer the question of how well a pre-trained multilingual model can generalize to an unseen language, and can equally be applied on the source and on the target-side.
The latter two experiments present two different methods for adaptation on $\newlang$ monolingual data, and focus on adding $\newlang$ on the target-side.

\subsubsection{Blind Decoding}%
\label{approach:blind_decoding}
\begin{figure}[tpb]
    \centering
    \includegraphics[width=1.0\linewidth]{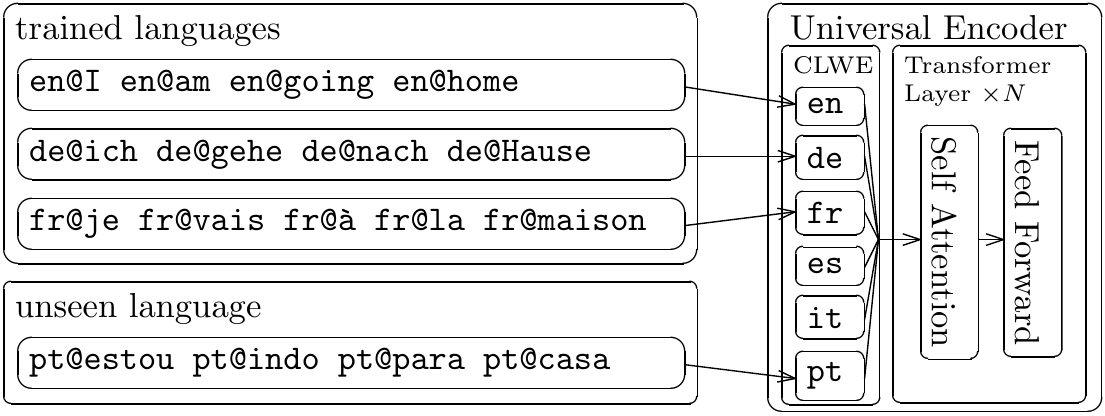}
    \caption{In \emph{blind decoding} we simply decode from an unseen language $\newlang$ without prior exposure to $\newlang$ whatsoever.
    The cross-lingual word embeddings (CLWE) serve as lookup table for $\newlang$ words.}%
    \label{fig:blind_decoding}
\end{figure}
To provide an indication of how well the universal encoder generalizes to unseen languages we let our multilingual model decode from $\newlang$ sentences, without any sorts of prior exposure to $\newlang$.
We call this method \emph{blind} decoding.
As depicted in Figure~\ref{fig:blind_decoding} we therefore employ the cross-lingual embeddings to lookup the $\newlang$ word vectors and then simply decode from the resulting sequence.
While we are also concerned about testing the NMT system ability to blindly decode \emph{to} a new language, we consider the target-side the less interesting side in this experiment, as the decoder couldn't possibly know the sentence level features, such as the $\newlang$ word order.
We anticipated that the language production ability should be very limited, while the language comprehension ability should -- through the information supplied by the word embeddings -- only be constrained to a lesser extend.
While the encoder may not know the syntax of an unseen language either, many syntactical concepts in language -- such as grammatical case, part-of-speech, grammatical genders, or tense -- are encoded at the word level\footnote{Take German for example, a noun would have different surface forms for different grammatical cases, which the fastText model should take into account and appropriately assign different word vectors.}.
As such we hypothesize that the word vectors trained on the new language alone provide enough syntactic level information to perform the language comprehension task to a certain degree.

\subsubsection{Learning Sentence-Level Features via Autoencoding}%
\label{approach:autoencoding}
As our first adaptation method towards $\newlang$ we propose two different ways of using our NMT model as an autoencoder, e.g. train to translate from $\newlang$ to $\newlang$.
The primary objective in doing so is to train the decoder $\newlang$ language model while maintaining the cross-lingual signal received from the encoder.

\paragraph{Denoising autoencoder}%
\label{approach:denoising_autoencoder}
For the first autoencoding approach we want to focus on teaching the model the target language syntax.
To achieve this, we train the model to perform denoising autoencoding:
the model gets noisy input sentences and is tasked with reconstructing the original, grammatically correct version of this sentence.
One major aspect of language syntax which sets apart different languages from each other is the word order.
By ``slightly'' shuffling the input sentence \citep{lample2017unsupervised}, e.g.\ displacing every word by up to $n$ positions from the original position, we can easily force the model to learn the correct word order.

\paragraph{Plain autoencoder}%
\label{approach:plain_autoencoder}
As our second autoencoding\footnote{due to the differing nature of the task we consider auto\emph{decoding} might be a more appropriate name} method we try training raw $\newlang{\rightarrow}\newlang$ translation while freezing the encoder parameters.
With this method we rely on the ability of our encoder to encode sentences from $\newlang$ in a suitable manner, as we directly take the encoder output for $\newlang$ sentences and learn to generate $\newlang$ sentences from it.
In order for the model to not just learn to copy words from the input sentence to the output sentence, we rely on the encoder output to be language independent enough to not retain any word level information from the source sentence.
This method thus tests the raw ability of our NMT system to learn the $\newlang$ language generation task conditioned on the encoder output $c$: $\mathbb{P}(Y_{\newlang} = y \mid H_{\text{enc}} = c)$.
In worst case this method trains the language generation ability conditioned on white noise.

\subsubsection{Backtranslation}%
\label{approach:backtranslation}
Lastly we would like to combine our model's ability to translate from $\newlang$ and its ability to adapt to $\newlang$.
We therefore use the in unsupervised NMT commonly employed approach of training on \emph{backtranslated} data \citep{sennrich-etal-2016-improving}.
Therefore one can use a joint NMT system translating between two languages in both directions to iteratively produce better translations, to then adapt the bootstrapped model on the improved synthetic generated data \citep{lample2017unsupervised}.
In our approach to backtranslation -- exploiting the model's ability to perform blind decoding -- we choose a simplified approach:
we perform one single round of backtranslation in which we use monolingual $\newlang$ data to
\begin{enumerate*}[label=\arabic*)]
    \setlength{\itemsep}{0pt}
    \item blindly decode from $\newlang$ to each one of our base languages $\lang \in \baselangs$ in order to generate synthetic parallel data, and then
    \item reversing the translation direction to train our model on this synthetic parallel data to translate from $\lang$ to $\newlang$.
\end{enumerate*}


\section{General Experimental Setup}%
\label{sec:experimental_setup}
\paragraph{Cross-lingual word embeddings}
As the basis for all of our monolingual word embeddings we use the pre-trained \emph{fastText} models provided on the fastText website\footnote{\url{https://fasttext.cc/docs/en/crawl-vectors.html}}.
These models provide 300-dimensional high-quality embeddings for full word units, accumulated on large web crawled monolingual text corpora.
Using the pivot approach described in section \ref{approach:cross_lingual_word_embeddings} we align these embeddings into a common embedding space using direct supervised optimization on the RCSLS criterion \citep{joulin-etal-2018-loss} while using the bilingual dictionaries provided by the MUSE repository\footnote{\citep[\url{https://github.com/facebookresearch/MUSE}]{conneau2017word}}.
In order to reduce the vocabulary size, we subsequently regenerate our vocabulary and the corresponding embedding vectors using only the words in our NMT training corpus.
Alignment accuracies and vocabulary sizes are listed in Table \ref{tab:vocab_size}.
\paragraph{Full word-units}%
\label{par:full_word_units}
\begin{table}[!tpb]
    \centering
    \begin{tabular}{c|rr}
        \hline
        lng   & vocab size & CLWE acc \tabularnewline
        \hline
        en    &  47,067 & 1.000 \tabularnewline
        de    &  98,747 & 0.554 \tabularnewline
        es    &  74,267 & 0.750 \tabularnewline
        fr    &  60,416 & 0.732 \tabularnewline
        it    &  74,962 & 0.696 \tabularnewline
        \hline
        total & 355,459 &       \tabularnewline
        \hline
        pt    &  67,237 & 0.734 \tabularnewline
        ru    & 166,651 & 0.627 \tabularnewline
        \hline
    \end{tabular}
    \caption{The number of words in the vocabulary generated from the training corpus for each of the languages.
             \emph{total} represents the size of the shared vocabulary of the base model.
             The column \emph{CLWE acc} describes the nearest neighbour accuracies for the cross-lingual embedding alignments to $\lang_{\text{pivot}} = \mathit{en}$.}
    \label{tab:vocab_size}
\end{table}
While using a joint BPE lets us share information between vocabularies and significantly reduces the size of the multilingual vocabulary, for our approach we choose to work with full word units, since our preliminary experiments have shown them to outperform the BPE models when used in combination with cross-lingual embeddings.
Furthermore, this gives us better compatibility with distant languages with separate alphabets and to avoids the problem of having to extend the learned BPE codes post-training when we add a new language to the model.
To alleviate the OOV word problem we use \emph{fastText} as the basis for our monolingual word embeddings, which, by using subword level information, gives us the ability to map unseen words into the embedding space.

For our multilingual vocabulary we merge our monolingual vocabularies and concatenate their embedding matrices, while dealing with duplicate words across different languages by encoding each word with a language specific prefix, e.g.\ we encode the English word \emph{bank} as \emph{\textbf{en@}bank}.
This also allows us to easily match any token in the vocabulary to its corresponding language, which is not trivial in a model with BPE units that are shared across multiple languages.
In inference we thus restrict the target vocabulary to the tokens matching the language we decode to, thereby forcing the model to only choose tokens from a single language in a situation where it would otherwise not do so, e.g.\ to decode to a yet entirely unseen language (see section \ref{sub:blind_decoding}).
This filtering furthermore provides a speedup in inference, as well as an improvement by around +0.4 BLEU in our basesystem evaluation.

\paragraph{Datasets}%
\label{par:datasets}
As our training, development and evaluation data sets we use multilingual transcriptions of TED talks \citep{cettoloEtAl:EAMT2012}.
For our basesystems we include English, German, Spanish, French and Italian as the base languages.
We train the basesystem on a total of 20 language pairs, and a parallel corpus size of $3,251,582$ sentences.
For our transfer learning experiments we use monolingual TED data for Portuguese and Russian, and parallel test data for BLEU score evaluation.
The monolingual data corpora contain $148,321$ sentences of Portuguese data and $187,843$ sentences of Russian data.
For test and development data we use the IWSLT dev2010 and tst2010 data sets.
All data is first tokenized and truecased with standard Moses tools.

\paragraph{NMT model}%
\label{evaluation:nmt_model}%
For our NMT model we use a Transformer model \citep{vaswani2017} with relative position encodings \citep{shaw-etal-2018-self}.
In accordance with our 300-dimensional fastText embeddings we use a model size of 300, and inner size of 1200, 6 attention heads and 6 layers.
Our implementation is based on the repository provided by \citet{kumar2018}\footnote{\url{https://github.com/Sachin19/seq2seq-con}}, which in turn is based on \emph{OpenNMT-py} \citep{klein-etal-2017-opennmt}.
We share the multilingual vocabulary, as well as the embedding parameters across the encoder and the decoder, and use tied output embeddings.
Refer to Appendix \ref{sec:training_parameters} for the full set of training parameters.

The evaluation scores for the multilingual basemodel, which we train on our available parallel data are listed in Table \ref{tab:basemodel_comparison}.
The scores represent the average over all of the 20 language pairs.
For the individual scores refer to Appendix B.
The \emph{frozen softmax} model serves as the main model for our transfer learning experiments.
It uses the cross-lingual embeddings in a pure lookup fashion, never updating the word vectors in NMT training.
This freezing ensures that the word embeddings stay in the shared embedding space, and we can as such trivially add new words to our vocabulary and align new languages into the shared embedding space.
The regular non-frozen softmax model serves for comparison with the frozen softmax model, as we do not know how well the pre-trained fastText embeddings fit the translation task.
We initialize this model to the trained cross-lingual embeddings while in training adapting them to the training data.
We can see that the frozen softmax model performs on par with its non-frozen counterpart, which suggests that the cross-lingual embeddings are very well suited for the translation task, despite them being trained on an unrelated task on purely monolingual data. \\
For the \emph{continuous output} model we employ the approach proposed by \citet{kumar2018}, which,
instead of calculating a multinomial distribution over a closed vocabulary, optimizes the network output directly towards the semantic information encoded in our pre-trained word embeddings.
This is ideal for our use case, since this is more suited towards extending the vocabulary.
Furthermore it helps us deal with our large multilingual vocabulary, as computational complexity and memory complexity in a continuous output model are independent of its vocabulary size.
While the model delivers inferior performance considering raw BLEU scores ony, training times amount to around one fifth of the training times for the softmax models. \\
Finally, in order to test the viability of our basesystem -- namely a full word-unit translation system with pre-trained word embeddings -- we also train a standard subword-unit translation system.
This model uses BPE trained on the multilingual corpus, reducing the vocabulary size from the $355,459$ words in our full word-unit vocabulary to $36,150$ BPE units.

\begin{table}[tpb]
    \centering
    \begin{tabular}[]{c|rr|rr}
        \toprule
        model & dev & $\Delta$ & test & $\Delta$ \tabularnewline
        \midrule
        BPE baseline   & 26.10 &       & 27.73 &       \tabularnewline
        \hline
        softmax        & 24.56 & -1.54 & 26.65 & -1.08 \tabularnewline
        frozen softmax & 24.96 & -1.14 & 26.60 & -1.13 \tabularnewline
        continuous     & 22.27 & -3.83 & 24.11 & -3.62 \tabularnewline
        \bottomrule
    \end{tabular}
    \caption{Comparison of average BLEU scores for the different variants of the multilingual base system}
    \label{tab:basemodel_comparison}
\end{table}

\section{Adding a New Language}%
\label{sec:language_addition}
\begin{table*}[!htpb]
    \centering
    \begin{tabular}{c|rrrrr|rr||rr}
        \toprule
        source-side method & pt-de & pt-en & pt-es & pt-fr & pt-it & ru-de & ru-en & $\varnothing$\,pt & $\varnothing$\,ru \tabularnewline
        \midrule
        blind decoding  & 20.6 & 36.4 & 30.7 & 29.3 & 24.4 & 11.1 & 12.8 & 28.3 & 11.9 \tabularnewline
        supervised      & 22.2 & 39.7 & 33.0 & 31.9 & 26.6 & 15.2 & 20.9 & 30.7 & 18.1 \tabularnewline
        \toprule
        target-side method & de-pt & en-pt & es-pt & fr-pt & it-pt & de-ru & en-ru & $\varnothing$\,pt & $\varnothing$\,ru \tabularnewline
        \midrule
        blind decoding  &  8.4 & 16.2 & 13.3 & 11.6 & 10.7 &  1.1 &  1.7 & 12.0 &  1.4 \tabularnewline
        autoencoder     & 17.0 & 28.1 & 27.1 & 21.5 & 20.8 &  8.1 &  8.7 & 22.9 &  8.4 \tabularnewline
        backtranslation & 21.3 & 34.6 & 32.3 & 26.4 & 25.4 & 13.6 & 13.9 & 28.0 & 13.8 \tabularnewline
        supervised      & 21.9 & 35.8 & 32.9 & 27.1 & 26.2 & 15.1 & 17.2 & 28.8 & 16.2 \tabularnewline
        \bottomrule
    \end{tabular}
    \caption{BLEU scores for decoding from (top) and to (bottom) Portuguese and Russian as the new language, with and without adaptation through various methods.}
    \label{tab:newlang_results}
\end{table*}
In the following we present our results for extending the multilingual basesystem by Portuguese and Russian.
Unless stated otherwise, we evaluate our methods on the section \ref{evaluation:nmt_model} \emph{frozen softmax} model.
For our adaptation methods we select the best performing model based on the development data BLEU scores.
\subsection{Blind Decoding}%
\label{sub:blind_decoding}
As our first experiment we investigate decoding to $\newlang$ without any sort of additional training or adaptation.
We therefore simply swap out the source or target-side embeddings and decode from or to $\newlang$.
As seen in Table \ref{tab:newlang_results} we can achieve BLEU scores of up to 36.4 with Portuguese as the source language.
While the average of 28.3 BLEU on the test set is significantly lower, it is still a considerably reasonable score, considering the model has never seen even a single Portuguese sentence.
On a more distant source language such as Russian, we achieve much lower scores.
Despite of the fact that the basesystem has never even seen a single sentence from any Slavic language the results are, however, still intelligible, achieving scores of up to 12.8 BLEU.

When decoding with $\newlang$ on the target-side, to force the model to decode to $\newlang$ as opposed to any $\lang \in \baselangs$ we need to filter out any target vocabulary words not associated with $\newlang$.
Initially, using our frozen softmax model to decode to Portuguese results in an average BLEU score of 3.0 BLEU.
For the continuous output model this amounts to 6.2 BLEU.
We suspect the main cause for the inferior softmax performance to be the missing bias for the new language, which the continuous output system does not require.
In fact, the models often get stuck in a decoding loop, which we combat by adding a postprocessing step remove the resulting duplicate words.
For both models this doubles the BLEU score, achieving up to even 16.2 BLEU on English-Portuguese.

\subsection{(Denoising) Autoencoder}%
\label{sub:denoising_autoencoder}
For our antoencoding methods we adapt on monolingual TED data using unchanged training parameters from the basesystem training.
When training to denoise via word order shuffling, we found -- in accordance with \citet{lample2017unsupervised} -- that displacing every word by up to $n = 3$ positions yields the best results.
We furthermore experiment with word dropout as additional noise, however, we did not observe any noticeable improvement over pure reordering noise.
Table \ref{tab:adaptation_comparison} shows the results for afterwards decoding to Portuguese and to Russian.
The denoising autoencoding results in slightly lower scores than the frozen encoder variant.
While the resulting scores clearly demonstrate the model ability to adapt to the new language -- even without noisy input -- we find the Russian scores to be disappointingly low.
Using both methods in conjunction with our postprocessing step (see section \ref{sub:blind_decoding}), we bring the translation to Russian up to a maximum of 8.7 BLEU and translation to Portuguese up to a maximum of 28.1 BLEU.
Table \ref{tab:newlang_results} shows the resulting BLEU scores for the individual language pairs.
The Portuguese and Russian softmax models both train for 1,000 iterations, completing after roughly 60 minutes on a \emph{GTX 1080 Ti} GPU.

We believe the significant improvement for the frozen encoder approach when adding noise to be an indication that the encoder representation still retains certain language specific word-level information.
In this case the noiseless model might find it easier to reconstruct the original sentence, thus learning less about the new language syntax in the process.
While this is an undesirable quality, this is to be expected considering just the fact that the encoder output sequence is equal in length to the input sentence.
\begin{table}[!pb]
    \centering
    \begin{tabular}{lrr}
        \toprule
        method & $\varnothing$\,pt & $\varnothing$\,ru \tabularnewline
        \midrule
        denoising autoencoder & 20.8 & 6.5 \tabularnewline
        frozen encoder        & 21.1 & 6.6 \tabularnewline
        + denoise             & 22.9 & 7.9 \tabularnewline
        ~~~+ postprocessing   & 22.9 & 8.4 \tabularnewline
        \midrule
        backtranslation       & 27.4 & 13.2 \tabularnewline
        + frozen encoder      & 28.0 & 13.8 \tabularnewline
        supervised baseline   & 28.8 & 16.2 \tabularnewline
        \bottomrule
    \end{tabular}
    \caption{A comparison of average test scores for different variations of the autoencoder methods (top), as well as the backtranslation method (bottom).}
    \label{tab:adaptation_comparison}
\end{table}
\subsection{Backtranslation}%
\label{sub:backtranslation}
Without any additional steps, our blind decoding approach has already enabled our basesystem to produce translations of up to 36.4 BLEU from Portuguese.
In our final experiment we exploit this for backtranslation to benefit the more difficult direction.
We train the Portuguese and the Russian model on backtranslated data for 6,000 and 10,000 iterations respectively, again with unchanged training parameters from basesystem training.
Table \ref{tab:newlang_results} shows the translation scores after training on backtranslated data for each of the language pairs.
Translation scores reach up to 34.6 BLEU, namely on English-Portuguese.
From the comparison in Table \ref{tab:adaptation_comparison} we can also see that the Portuguese backtranslation scores are almost on par with our supervised model trained on bilingual data, falling short by just 0.8 BLEU.
Again, freezing the encoder in training results in improved translation, albeit to a lesser extent than for the autoencoding approach.
We note that training the Russian system on $en{\rightarrow}ru$ data only also improves the performance for $de{\rightarrow}ru$, in our experiments sometimes even outperforming the language pair it is trained on.
We see this as an indication that this training mainly affects the decoder language model for Russian.
This furthermore means that the decoder, when decoding to Russian, is able to use a similar amount of information from the encoder representation of the source sentence.
This either means that the encoder representation of English and German are very similar in quality, being able to extract the same amount of meaning.
Alternatively it means that the decoder does not learn to use the full amount of information encoded into the representation of the English source sentence, meaning the decoder is lacking in translation adequacy while leaning towards better fluency.


\section{Conclusion}
In this work we have looked into adding a new language to a previously trained multilingual NMT system in an unsupervised fashion.
We explored the possibility of reusing the existing latent sentence representation and adding a language by adapting the decoder on monolingual data only.
As part of our approach we show the feasability of doing multilingual NMT with pre-trained cross-lingual word embeddings.
By blindly decoding from Portuguese, e.g.\ translating without any prior exposure towards Portuguese aside from the cross-lingual word embeddings, using a basesystem containing multiple Romance languages we achieve scores of up to 36.4 BLEU for Portuguese-English, thereby demonstrating the universal encoder ability to generalize to an entirely unseen language.
While this works significantly worse on a more distant language, this model -- which has never seen even a single sentence from any Slavic language -- is still able to achieve up to 13 BLEU for Russian-English.
Furthermore, by applying blind decoding on the target-side we have been able to achieve up to 16.2 BLEU when decoding \emph{to} Portuguese.
To this end we looked into the recently proposed continuous output representation approach.

In an attempt to train the mapping from our sentence representation to a new target language we use our model as an autoencoder.
By training to translate from Portuguese to Portuguese while freezing the encoder we have achieved up to 26 BLEU for English-Portuguese.
Adding artificial noise to the input to let the model learn the correct word order gains us an additional 2 BLEU.
Translating to Russian we achieve up to 8.7 BLEU for English-Russian.

Lastly we have explored a more practical approach of learning the new language by training the system on backtranslated data.
To this end we exploit our model's ability to produce high quality translations from an unseen source-side language to generate the synthetic data.
The training on the synthetic data has yielded up to 34.6 BLEU, again on English-Portuguese, attaining near parity with a model trained on real bilingual data.
Translating to Russian yields at most 13.9 BLEU for English-Russian.
Considering the low English-Russian baseline score of 17.2 BLEU we suspect the overall low Russian scores to partly be an issue with the low quality word embedding alignment.

\subsection{Future Work}
In the future we would like to explore the question of how the composition and number of languages in the base system affects the ability to perform the transfer learning, and whether seeing a wide variety of different languages will help the translation system in generalizing to new languages -- especially more distant ones.
In order to further improve the performance on source-side unseen languages we would like to explore various methods for the adaptation of the encoder.
While the focus of this work lies mainly on adapting the decoder, an improvement on the source-side would especially benefit the backtranslation results for more distant languages.
As an alternative to teaching the new language to the decoder via autoencoding, we would like to explore the possibility of using generative adversarial training on top of a translation system with a continuous output representation.

\section{Acknowledgement}
The projects on which this paper is based were funded by the Federal Ministry of Education and Research (BMBF) of Germany under the numbers 01IS18040A.
The authors are responsible for the content of this publication.

\bibliography{anthology,main}
\bibliographystyle{acl_natbib}

\clearpage
\appendix

\section{Training Parameters}
\label{sec:training_parameters}
\begin{lstlisting}
general settings:
  layers 6
  rnn_size 300
  word_vec_size 300
  transformer_ff 1200
  heads 6
  warmup_init_lr 1e-8
  warmup_end_lr 0.0007
  min_lr 1e-9
  encoder_type transformer
  decoder_type transformer
  position_encoding
  max_generator_batches 2
  param_init_glorot
  label_smoothing 0.1
  param_init 0
  share_embeddings
  share_decoder_embeddings
  generator_layer_norm
  warmup_steps 4000
  learning_rate 1
\end{lstlisting}
\begin{lstlisting}
vmf model:
  dropout 0.1
  batch_size 8192
  batch_type tokens
  normalization tokens
  accum_count 2
  optim radam
  adam_beta2 0.9995
  decay_method linear
  weight_decay 0.00001
  max_grad_norm 5.0
  lambda_vmf 0.2
  generator_function continuous-linear
  loss nllvmf
\end{lstlisting}
\begin{lstlisting}
softmax model:
  dropout 0.2
  batch_size 1536
  batch_type tokens
  accum_count 6
  optim adam
  adam_beta2 0.999
  decay_method noam
  max_grad_norm 25
\end{lstlisting}

\newpage
\section{Multilingual Basesystem Scores}
\label{appendix:basesystem_scores}
\begin{table}[htpb]
    \centering
    \begin{tabular}[]{|c|*{5}{r}|}
        \hline
        $\lang$ & \texttt{de} & \texttt{en} & \texttt{es} & \texttt{fr} & \texttt{it} \tabularnewline
        \hline
        \texttt{de} &      & 30.6 & 21.4 & 23.0 & 18.5 \tabularnewline
        \texttt{en} & 25.9 &      & 34.6 & 33.1 & 26.5 \tabularnewline
        \texttt{es} & 21.8 & 40.8 &      & 29.8 & 25.1 \tabularnewline
        \texttt{fr} & 19.1 & 32.5 & 25.3 &      & 22.8 \tabularnewline
        \texttt{it} & 18.9 & 30.2 & 25.7 & 26.7 &      \tabularnewline
        \hline
    \end{tabular}
    \caption{Individual language pair test scores for the section \ref{evaluation:nmt_model} \emph{frozen softmax} basesystem.
             Lines represent the source language, while columns represent the target language.}
    \label{tab:frozen_softmax}
\end{table}

\end{document}